\pdfoutput=1

\documentclass[11pt]{article}

\usepackage[final]{acl}

\usepackage{times}
\usepackage{latexsym}

\usepackage[T1]{fontenc}

\usepackage[utf8]{inputenc}

\usepackage{microtype}

\usepackage{inconsolata}

\usepackage{graphicx}
\usepackage{booktabs}
\usepackage{paralist}

\renewenvironment{enumerate}[1]{\begin{compactenum}#1}{\end{compactenum}}
\usepackage{tabularx}
\usepackage{comment}
\usepackage{xcolor}
\usepackage{soul}

\usepackage{tikz}

\usepackage{textcomp}

\title{Greenback Bears and Fiscal Hawks:\\Finance is a Jungle and Text Embeddings Must Adapt}

\author{Peter Anderson \enspace Mano Vikash Janardhanan \enspace Jason He \enspace Wei Cheng \enspace Charlie Flanagan\\
  Balyasny Asset Management \\
  \small{\texttt{\{peanderson, mjanardhanan, jhe, wcheng, cflanagan\}@bamfunds.com}}}

\begin{document}
\maketitle

\begin{abstract}
Financial documents are filled with specialized terminology, arcane jargon, and curious acronyms that pose challenges for general-purpose text embeddings. Yet, few text embeddings specialized for finance have been reported in the literature, perhaps in part due to a lack of public datasets and benchmarks. We present BAM embeddings, a set of text embeddings finetuned on a carefully constructed dataset of 14.3M query-passage pairs. 
Demonstrating the benefits of domain-specific training, BAM embeddings achieve Recall@1 of 62.8\% on a held-out test set, vs. only 39.2\% for the best general-purpose text embedding from OpenAI. Further, BAM embeddings increase question answering accuracy by 8\% on FinanceBench and show increased sensitivity to the finance-specific elements that are found in detailed, forward-looking and company and date-specific queries. To support further research we describe our approach in detail, quantify the importance of hard negative mining and dataset scale.

\end{abstract}

\section{Introduction}

Portfolio managers and analysts have access to millions of financial documents. Text embeddings are a key component of the information retrieval and retrieval-augmented generation (RAG) systems \cite{lewis2020retrieval} that can help extract insights from this mass of information. However, the financial domain poses unique challenges for text embeddings. Financial documents are filled with specialized terminology (`par value', `stagflation'), jargon (`Chinese wall'), curious acronyms (`CAGR', `DCF', `VIX'), and technical terms and company names that collide with ordinary words (`short', `forward', `spread', `Apple', `Stripe'). Is a CDO similar to a CFO? And what is a Greenback bear\footnote{A Greenback bear is an investor who believes the US dollar will decline in value. A fiscal hawk argues for a reduction in government spending.}?

Despite their importance, few text embeddings specialized for finance have been reported in the literature. 
To address this gap, we present BAM embeddings, a set of 
text embeddings optimized for financial document retrieval. 
BAM embeddings are based on Multilingual-E5 \cite{wang2024multilingual}, further finetuned on a carefully filtered, clean dataset of 14.3M query-passage pairs (6B tokens) constructed from 2.8M financial documents.
While we cannot release our dataset, we describe in detail our data curation and query generation strategy, finetuning process, and approach to deployment in a high-priority application.

On a held-out set of 447K query-passage pairs, BAM embeddings achieve Recall@1 of 62.8\%, far surpassing the Multilingual-E5 base model (34.3\%) as well as large closed-source models (e.g., OpenAI's 
3072-dim text-embedding-3-large model, 39.2\%). Quantitatively, we show that hard negative mining (+5.3\%) and data scale (+4.5\%) are critical to achieving this performance. Deploying BAM embeddings in an application alongside traditional lexical search (Okapi BM25), we find that BAM embeddings outperform lexical search over all query lengths. Notably, vector search with BAM embeddings improves as queries become longer and more detailed, while lexical search degrades.

Finally, we evaluate BAM embeddings in a public RAG benchmark using FinanceBench \cite{islam2023financebench}. Replacing OpenAI's ada-002 embeddings with ours increases question answering correctness by 8\%.  
Qualitatively, we observe that after finetuning, embeddings are more sensitive to company names, tickers and financial metrics, leading to improved performance detailed, forward-looking, and company or date-specific queries.

\section{Related Work}

\paragraph{Financial LLMs}

Much of the previous work in Financial NLP has focused on adapting large language models (LLMs) to finance. FinBERT \cite{yang2020finbertpretrainedlanguagemodel} was the first LLM to demonstrate the benefits of pretraining on company reports, broker research and earnings transcripts. When finetuned for financial sentiment classification, FinBERT outperformed the standard BERT model pretrained on generic text \cite{devlin2019bertpretrainingdeepbidirectional}. FLANG \cite{shah-etal-2022-flue} demonstrated that the gains from pretraining on financial documents extended to other downstream tasks, including news headline classification, named entity recognition (NER), structure boundary detection, and question answering in the finance domain. Compared to these 100M parameter models, the 50B parameter BloombergGPT model \cite{wu2023bloomberggptlargelanguagemodel} facilitated evaluation via few-shot prompting rather than finetuning, again outperforming general-purpose LLMs. Continuing the trend towards chat-based approaches, PIXIU \cite{xie2023pixiu} developed a financial instruction dataset and benchmark generalizing 9 financial NLP tasks, and used it to finetune LLaMA \cite{touvron2023llamaopenefficientfoundation}.

\paragraph{Domain-Specific Text Embeddings}

Even with pretraining or finetuning for finance, LLMs require up-to-date information, which is typically retrieved using text embeddings \cite{lewis2020retrieval}. Domain-specific embeddings have been well-studied in healthcare \cite{alsentzer-etal-2019-publicly, 10.1093/bioinformatics/btz682} and law \cite{chalkidis-etal-2020-legal}. Surprisingly, no text embeddings specialized for financial document retrieval have been reported in the literature, perhaps in part due to a lack of public datasets and benchmarks on which to train and evaluate. Standard benchmarks such as MTEB \cite{muennighoff2022mteb} contain no company reports, broker research or earnings transcripts\footnote{The most relevant MTEB datasets -- FIQA 2018 \cite{Maia2018WWW18OC} and Financial PhraseBank \cite{Malo2014GoodDO} -- are based on financial news and blog posts.}, while finance-oriented benchmarks such as FinanceBench \cite{islam2023financebench} and PIXIU \cite{xie2023pixiu} are designed to evaluate LLMs not text embeddings. We address the lack of finance text embeddings by releasing BAM embeddings. Similar to \citet{ma-etal-2021-zero, cho-etal-2022-query} and others, our approach relies on synthetic query generation for training data.

\section{Dataset}

To finetune and evaluate BAM embeddings, we construct a dataset of 15.2M \textit{query-passage} pairs. Text \textit{passages} are drawn from financial documents (refer Section \ref{sec:raw_documents}). For each passage, a matching \textit{query} is generated using a few-shot prompted LLM (refer Section \ref{sec:query_generation}).

\subsection{Sampling Text Passages}
\label{sec:raw_documents}

\paragraph{Raw Documents} Text passages are sourced from 2.8M documents published in the two-year period ending 31 March 2024.

\paragraph{Parsing and Splitting} Most documents are stored in PDF format. We convert them to text using an internal PDF-to-text conversion tool, then split the text into passages. Our splitting strategy recursively splits text based on a generic list of typical paragraph separators (`$\backslash$n$\backslash$n', `$\backslash$n', etc), while attempting to avoiding splitting questions and related answers (which are common in transcripts of company presentations). We choose a maximum passage length of 512 tokens (350--400 words). This typically provides sufficient context to understand the text while shorter passages can be difficult to interpret. We use regular expressions and heuristics to remove legal disclosures and most tables from the dataset (we focus on text passage retrieval and leave information extraction from tables to future work).

\paragraph{Document Context} After splitting, text passages are frequently missing crucial information such as the name and stock ticker of the company referenced in the text, and the date. We augment each text passage by prepending one line of document context. The content of this context line differs by document type --- for a company earnings transcript it contains the company name, ticker, and the event (e.g., `FY23 earnings call'). We use the same approach when finetuning the embeddings and in deployment.

\subsection{Query Generation}
\label{sec:query_generation}

\begin{table}
  \centering
  \small
  \begin{tabularx}{\columnwidth}{X}
    \toprule
    cotton corn production cost trends  \\
    wet wipes new products  \\
    sector rotation this week  \\
    Reliance CapEx  \\
    artificial intelligence at medtronicas  \\
    \scriptsize{How has brexit affected supply chain operations and companies?} \\
    Chinese real estate new residential sales  \\
    beazley price  \\
    What is going on with pulpwood costs?  \\
    \tiny{How is Teck Resources generating shareholder value through structured separation?}  \\
    \bottomrule
  \end{tabularx}
  \caption{Examples of the human-written queries used to seed synthetic queries via few-shot prompting.}
  \label{tab:few-shot-queries}
\end{table}

Rather than finetuning on user-generated queries, we choose to synthetically generate all queries in the dataset. Synthetic generation is highly scalable, eliminates privacy concerns, ensures complete coverage of all text passages in the corpus, and enables finetuning and evaluating on queries that are longer and more complex than user queries.

\paragraph{Few-Shot Examples} To seed our query generation strategy, we randomly sample text passages and ask a group of quantitative researchers, engineers and product managers to write a query for each passage. Annotators are instructed that the passage should be a top result for that query in a world-class document retrieval system (refer Appendix). Using this approach, we collect 231 passage-query pairs to use for few-shot prompting (see Table \ref{tab:few-shot-queries} for randomly selected examples).

\paragraph{LLM Query Generation}
We prompt an LLM to generate a single matching query for every text passage in the corpus, or output `SKIP' if a query can't be generated. To encourage quality and diversity, two passages with human-written queries are randomly selected and included as few-shot examples (refer to the Appendix for the complete prompt). Trading off computational cost and query quality, we use the Mistral 7B Instruct model \cite{jiang2023mistral} for query generation. Even using vLLM \cite{kwon2023efficient} for high throughput, several weeks of A100 gpu time are required to generate queries for the entire dataset. In initial experiments we also used a second LLM call to paraphrase the generated queries for increased diversity and challenge, but found this refinement was less effective than simply generating more queries.

\begin{figure*}[t]
  \includegraphics[width=\textwidth]{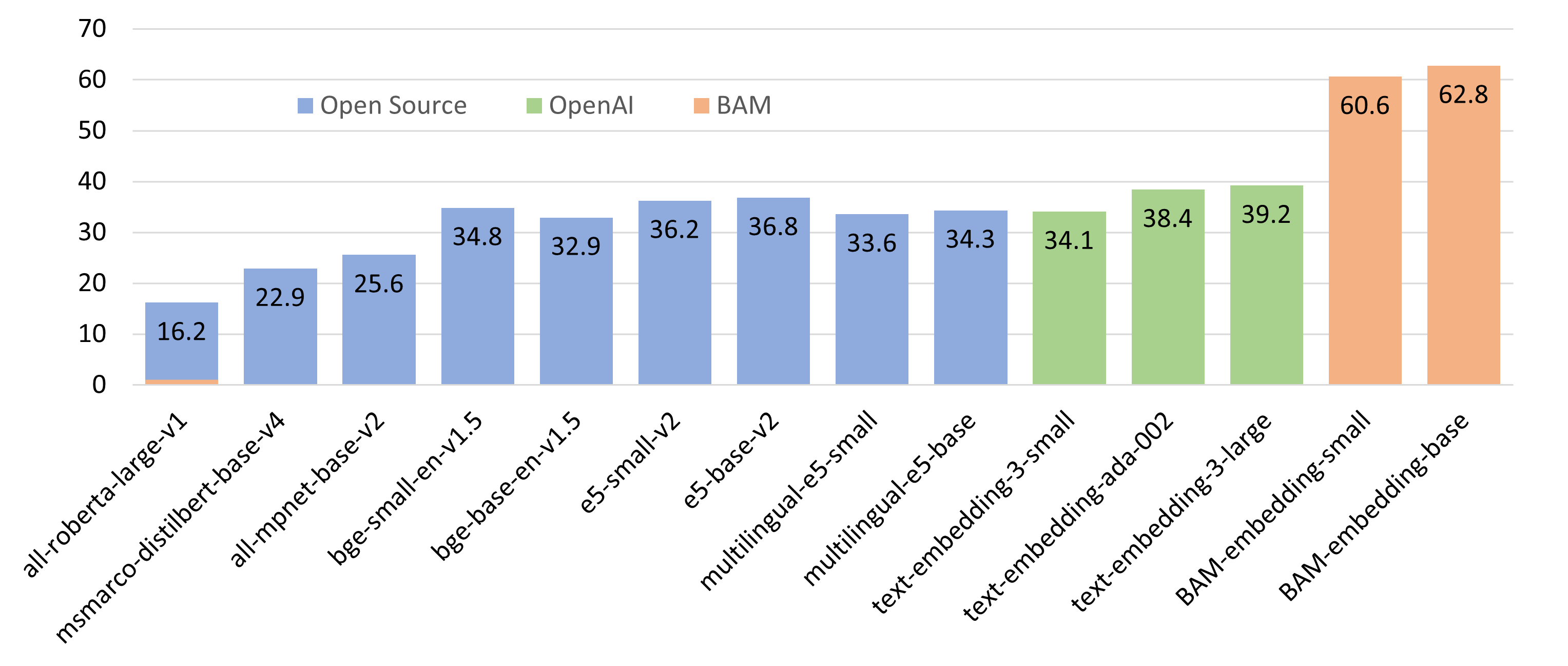}
  \caption{Passage retrieval results: Recall@1 on a held-out test split of 447K query-passage pairs. BAM embeddings finetuned for financial document retrieval significantly outperform general-purpose embeddings.}
  \label{fig:results}
\end{figure*}

\paragraph{Filtering}
Passages for which the LLM failed to generate a valid query are identified by the presence of the special `SKIP' token provided in the prompt, or by responses that begin with identified phrases that indicate failure such as `no query', `no question' or `understood'. These are removed from the dataset, along with all duplicate queries, which arise when multiple text passages generate the same query, e.g., `Apple 2024 EPS’. After filtering (including filtering performed during hard negative mining, refer Section \ref{sec:model}) the final dataset consists of 15.2M query-passage pairs, which are separated into train-val-test splits containing 14.3M, 444K, and 447K examples, respectively. All passages from the same document are assigned to the same split. This avoids contaminating the val or test splits if documents are repetitive.

\paragraph{Query Realism}
We do not aim to replicate the query distribution from our legacy document retrieval system, which are typically short, simple keyword queries. These are shaped by user interactions with a weaker BM25-based system. Instead, we aim to support longer, more complex queries, which we now encourage (refer Section \ref{sec:real}).

\section{BAM Embeddings}
\label{sec:model}

\paragraph{Baseline Model} We finetune the Multilingual-E5 model \cite{wang2024multilingual}, which is pretrained on 1B multilingual text pairs and pre-finetuned on a combination of general-purpose labeled datasets. Multilingual-E5 embeddings are based on the XLM-RoBERTa architecture \cite{XLM-RoBERTa} which has three sizes: \textit{small} (118M model params, 384-dim embeddings), \textit{base} (278M model params, 768-dim embeddings), and \textit{large} (560M model params, 1024-dim embeddings). We finetune small and base models; in preliminary experiments we found that the large model performed no better than the base model so we discontinued finetuning.

We select Multilingual-E5 because the model achieved competitive baseline retrieval performance on our dataset, and because the embedding is mean-pooled\footnote{Mean-pooling is applied to the output of the final transformer layer across all tokens in the query or passage.}. This enables us to calculate a contextualized embedding for any sentence within a passage by simply mean-pooling over the relevant subset of tokens. In our application, we exploit this affordance to highlight the sentences in a document that are most relevant to a user's query, based on the similarity between each sentence and the query. Embedding models based on the output of the CLS token, such as BGE \cite{bge_embedding}, do not offer this capability.

\paragraph{Finetuning} We use the standard InfoNCE contrastive loss \cite{oord2019representation} that requires the model to identify the positive passage for each query from a set of negative passages. Negative passages are comprised of in-batch negatives (the other passages in the same minibatch) plus 3 hard negatives per query (see below).

We finetune for 3 epochs using a batchsize of 512 and initial learning rate $\{3, 2, 1\} \times 10^{-5}$ for the \{small, base, large\} models. Following \citet{wang2024textembeddingsweaklysupervisedcontrastive}, during finetuning, evaluation and in deployment, we add the prefixes `query: ' and `passage: ' to queries and passages respectively, allowing the model to better represent short queries and long passages in the same embedding space.

\paragraph{Hard Negative Mining} Hard negative mining improves the quality of learned embeddings by introducing negative examples that are more challenging to detect than randomly-sampled negatives \cite{gao-etal-2021-simcse, karpukhin-etal-2020-dense}. For each query, we identify 3 hard negative passages using an early version of our model finetuned with 37\% of the full dataset and no hard negatives. Specifically, we embed all 15.2M queries and passages in the dataset, retrieve the top 1K passages for each query, and label the passages ranked 200--202 places lower than the positive passage as hard negatives. This hyperparameter was set after testing several different options in initial experiments. If the positive passage is not in the retrieved passages, the query-passage pair is removed from the dataset, as manual inspection indicates that these are typically low-quality pairs.

\begin{table*}[]
\small
\centering
\begin{tabularx}{\textwidth}{Xrcccccccc}
   \textbf{}                      &                    & \textbf{}    & \multicolumn{3}{c}{\textbf{Passage Retrieval}}                                                           & \textbf{} & \multicolumn{3}{c}{\textbf{Document Retrieval}}                                                          \\ \cmidrule(lr){4-6} \cmidrule(l){8-10} 
\textbf{Embedding}             & \textbf{Dim} && \multicolumn{1}{l}{\textbf{R@1}} & \multicolumn{1}{l}{\textbf{R@10}} & \multicolumn{1}{l}{\textbf{R@50}} & \textbf{} & \multicolumn{1}{l}{\textbf{R@1}} & \multicolumn{1}{l}{\textbf{R@10}} & \multicolumn{1}{l}{\textbf{R@50}} \\
\midrule
text-embedding-3-small  & 1536 && 34.1                         & 68.3                          & 84.4                          &  & 61.1                         & 85.0                          & 90.1                          \\
text-embedding-ada-002  & 1536 && 38.4                         & 71.6                          & 85.7                          &  & 64.3                         & 86.2                          & 91.3                          \\
text-embedding-3-large  & 3072 && 39.2                         & 73.7                          & 87.8                          &  & 64.9                         & 87.4                          & 91.8                          \\
\midrule
multilingual-e5-small & 384  && 33.6                         & 67.1                          & 82.1                          &  & 61.2                         & 84.2                          & 89.7                          \\
multilingual-e5-base  & 768  && 34.3                         & 68.2                          & 83.1                          &  & 62.6                         & 85.6                          & 91.1                          \\
\midrule
BAM-embedding-small     & 384  && 60.6                         & 89.6                          & 96.6                          &  & 81.4                         & 95.5                          & 97.9                          \\
BAM-embedding-base      & 768  && \textbf{62.8}                         & \textbf{91.0}                          & \textbf{97.3}                          &  & \textbf{83.0}                         & \textbf{96.3}                          & \textbf{98.4}                         \\
...1 hard negative, full training data  &  768  && 61.8                         & 90.4                          & \textbf{97.3}                          &  & 82.2                         & 96.1                          & \textbf{98.4}                         \\
...No hard negatives, full training data  &  768  && 57.5                       & 88.8                          & 97.2                          &  & 79.5                        & 95.8                          & 98.5                         \\
...No hard negatives, 37\% training data   &  768  && 53.0                         & 86.1                          & 96.1                          &  & 76.4                         & 94.6                          & 97.9                         \\
\bottomrule
\end{tabularx}
\caption{Base-sized BAM embeddings outperform the much larger OpenAI embeddings (top panel) and the baseline model (middle panel) on all recall metrics. Ablation studies (bottom 3 rows) highlight the importance of hard negative mining and scaling the finetuning data.}
\label{tab:recall}
\end{table*}

\section{Results and Analysis}

\subsection{Retrieval}

We evaluate the retrieval performance of BAM embeddings using the held-out test set of 447K query-passage pairs. We benchmark against 12 other models, including the E5 \cite{wang2024textembeddingsweaklysupervisedcontrastive} and BGE \cite{bge_embedding} families of models, and text embeddings from OpenAI. Since each query has only a single correct passage, we report Recall@$K$ with $K=1,10,50$ rather than NDCG. Since the users of our system (both human and LLM agent) have limited attention, we focus on Recall@1 although the same trends hold across other values of $K$.

As illustrated in Figure \ref{fig:results}, finetuning the 768-dim Multilingual-E5 base model improves Recall@1 from 34.3\% to 62.8\% -- surpassing an array of open-source and close-source models including OpenAI's 3072-dim text-embedding-3-large model, which achieves 39.2\% Recall@1 (with a much larger embedding). We observe similar gains with the small model, which achieves 60.6\% Recall@1.

\paragraph{Ablation Studies} In Table \ref{tab:recall} we report Recall@K for both passage retrieval and document retrieval (i.e., retrieving any passage from the document containing the correct passage). We compare to OpenAI embeddings (top panel) and the baseline models (middle panel). We find that base-sized BAM embeddings perform best on every metric, and hard negative mining and data scale are crucial to achieving this performance. 

Without hard negative mining, passage Recall@1 for the base-sized model drops from 62.8\% to 57.5\%, although 1 hard negative is sufficient to capture most of the benefits (61.8\% vs. 62.8\% with 3 hard negatives). Reducing the amount of training data to 37\% of the full dataset (representing only one year of data, and 4 document types instead of 7) further reduces Recall@1 from 57.5\% to 53.0\%, demonstrating the value of scaling the dataset over a large document corpus.

\paragraph{Qualitative Analysis} 
In Table \ref{tab:examples} we provide an example of how query similarity changes before and after finetuning. For more general insights, we randomly select 100 queries with a large improvement in recall under the finetuned model, and 100 queries with no improvement, and ask ChatGPT to identify qualitative differences between the queries. 

According to ChatGPT, the queries that improved the most are company-specific (focused on individual companies and particular quarters or fiscal years), forward-looking (referencing future projects and growth), searching for specific financial metrics (such as PE ratios or adjusted net income), and phrased as questions rather than statements. Queries about general financial and economic concepts, such as `capital issuance', improved the least.

\begin{table*}
  \centering
  \small
  \begin{tabularx}{\textwidth}{X X}
    \toprule
    \textbf{Query:} {\color{blue}ASMI}'s {\color{purple}price-to-earnings} ratio \\
    \midrule
    \textbf{Nearest Neighbors Before Finetuning:}  & \textbf{Nearest Neighbors After Finetuning:} \\
    1. {\color{blue}ASMI} - earnings report analysis  &   1. {\color{blue}ASMI} - earnings report analysis \\
    2. \st{ASMedia}'s share price            &   2. {\color{blue}ASMI} outlook \\
    3. {\color{blue}ASMI} revenue goals               &   3. What is the current market outlook for {\color{blue}ASMI}? \\
    4. \st{Alcoa}'s {\color{purple}price to earnings} ratio  &   4. What is the rating and {\color{purple}PE} ratio of {\color{blue}ASMI}? \\
    5. \st{Aviva}'s cost-to-income ratio     &   5. What is {\color{blue}ASMI}’s revenue growth forecast for 2024?\\
    6. \st{Axis Bank}'s cost-to-income ratio &   6. Interesting read on {\color{blue}ASMI} stock.\\
    \bottomrule
  \end{tabularx}
  \caption{Nearest neighbor queries before and after finetuning Multilingual-E5. Before finetuning, embeddings capture too much lexical similarity, e.g. ASMI is similar to ASMedia; price-to-earnings is similar to cost-to-income. After finetuning, embeddings are more sensitive to tickers and stock names.
 }
  \label{tab:examples}
\end{table*}

\subsection{Results on FinanceBench}

Given the lack of public benchmarks for financial document retrieval, in this section we report results on FinanceBench \cite{islam2023financebench}, a benchmark for financial question answering.
We co-opt FinanceBench's retrieval-augmented generation (RAG) setting to assess how retrieval with different text embeddings affects answer accuracy.

\begin{figure}[t]
  \includegraphics[width=\columnwidth]{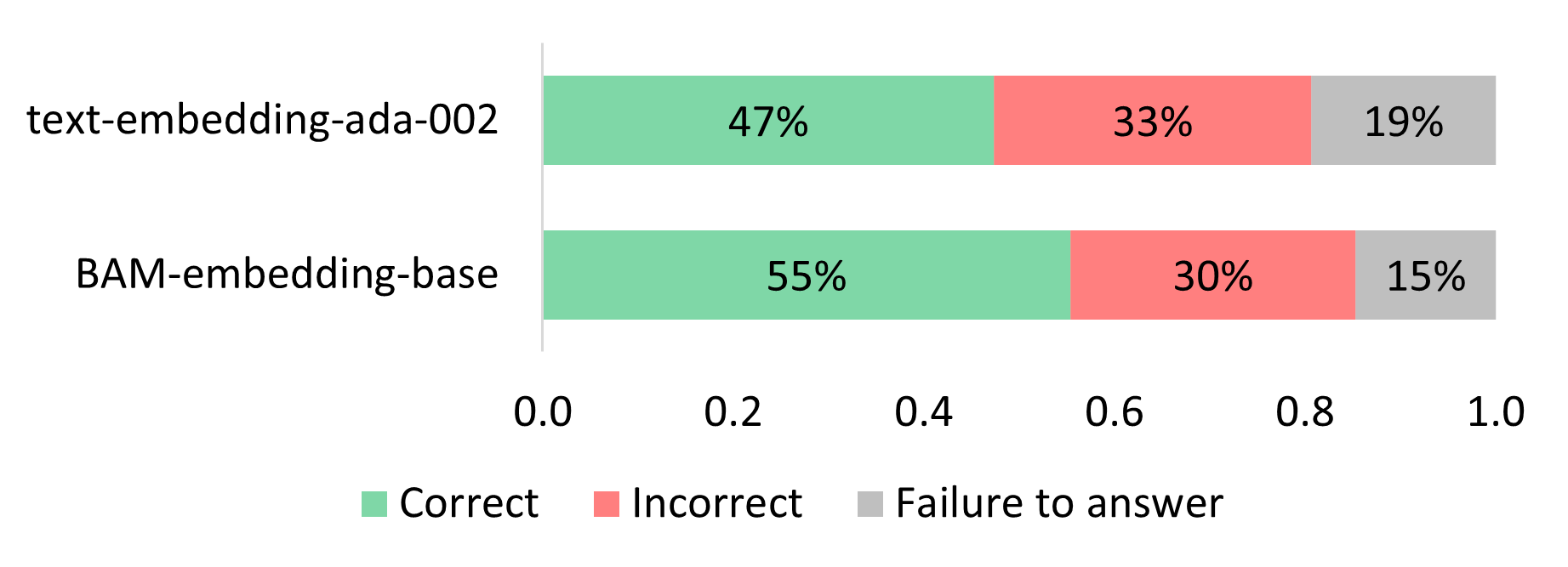}
  \caption{FinanceBench results under the Shared Vector Store setting. Replacing OpenAI ada-002 embeddings with BAM embeddings increases accuracy by 8\%. }
  \label{fig:financebench}
\end{figure}

We focus on the Shared Vector Store setting, in which text passages from a collection of 368 10-K and 10-Q reports are stored in a single vector database, which is queried by an LLM to answer 150 questions derived from those documents. 
\citet{islam2023financebench} report that
GPT-4 correctly answers only 19\% of questions using OpenAI ada-002 text embeddings. However, the original RAG pipeline has several weaknesses. We improve it by:
\begin{enumerate}
    \item Using an LLM to rewrite the question before querying the vector store. This eliminates distracting text containing formatting instructions
    \item Prepending the filename of the parent document to the beginning of each text passage, which preserves document context including the company name and the filing date
    \item Prompting the LLM to generate concise answers (which are more in keeping with the gold answers and easier to evaluate)
    \item Replacing GPT-4 with GPT-4o
\end{enumerate}
\noindent
Based on human evaluation, our improved RAG pipeline achieves 47\% accuracy vs. 19\% reported by \citet{islam2023financebench}. As illustrated in Figure \ref{fig:financebench}, replacing the 1536-dim ada-002 embeddings with 768-dim BAM embeddings improves accuracy further to 55\% -- even though most FinanceBench questions are table-based, and BAM embeddings were optimized for text passage retrieval. The remaining errors are mostly attributable to the LLM (extracting numbers from tables and calculating derived metrics such as operating margin).

\subsection{Real-world Deployment}
\label{sec:real}

\paragraph{Application} We have deployed BAM embeddings in a RAG service that indexes 5.7M financial documents (1.3TB of raw data), providing a backend API for 3 different frontend applications (two market intelligence and search platforms and a chatbot). We use OpenSearch because it supports approximate nearest neighbor vector search in conjunction with traditional filtering operations. Filters are used to restrict search results based on document date ranges, company tickers, and tags such as document type, event name, data vendor, etc. In addition to OpenSearch, we maintain a NoSQL database to store document fields and metadata that would add excessive overhead to OpenSearch.

\paragraph{Weight Averaging} Before deploying BAM embeddings, we average the parameters of 5 finetuned checkpoints (trained for between 2.5 and 3 epochs) with the baseline model, with a 50\% weighting on the baseline model and 10\% weighting on each checkpoint. We are motivated by \cite{wortsman2022robustfinetuningzeroshotmodels, pmlr-v162-wortsman22a} who show that averaging the weights of zero-shot and finetuned models improves accuracy and robustness to out of distribution queries. Robustness is an important consideration because we expect the distribution of user queries to drift over time (and may not perfectly match our generated queries to start with). Weight averaging sacrifices 1.1\% Recall@1 on our dataset, but improves NDCG@10 on a representative out-of-domain dataset (FiQA 2018) by 2.2\% compared to the final checkpoint.

\begin{figure}[t]
  \includegraphics[width=\columnwidth]{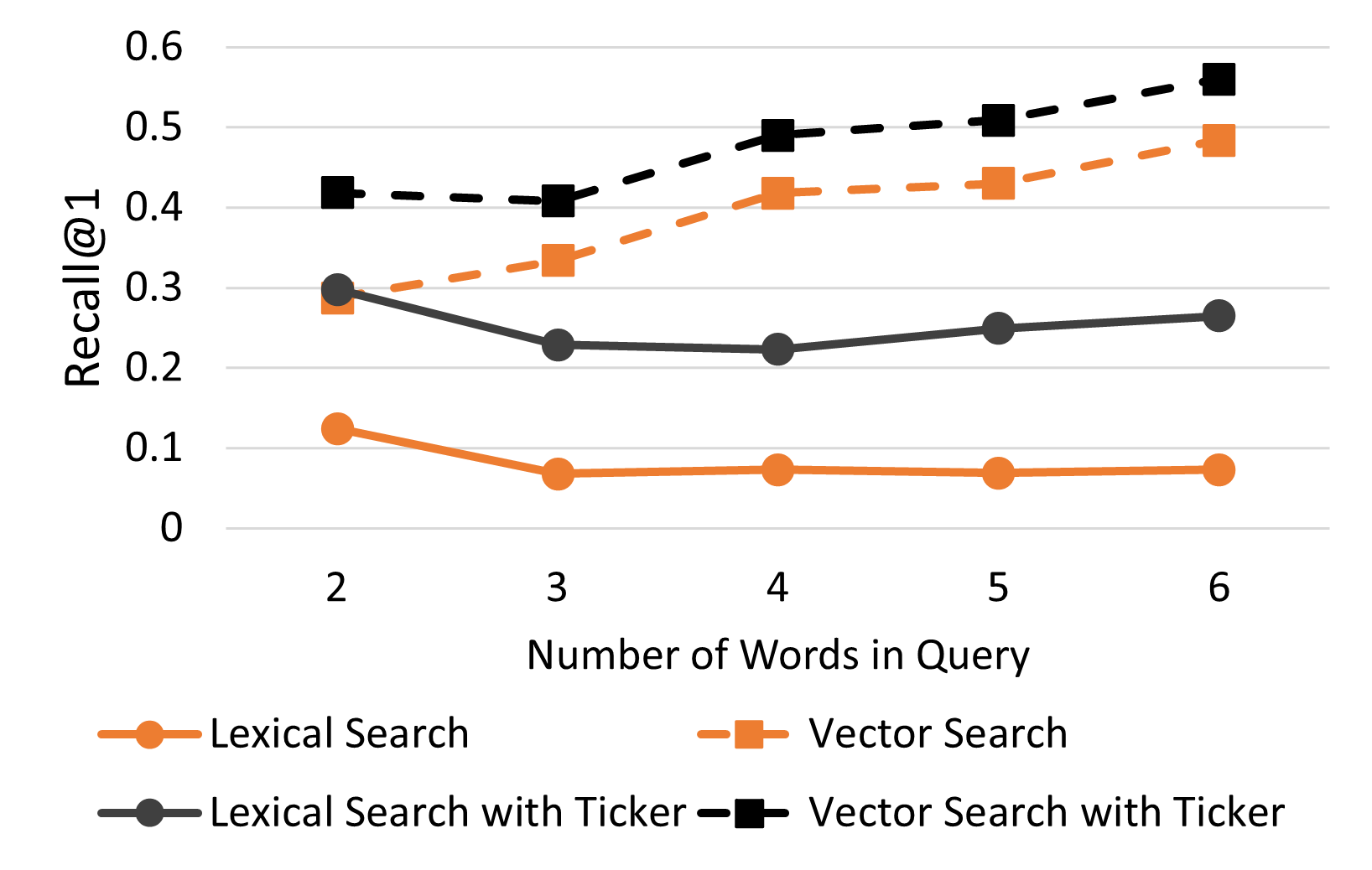}
  \caption{Comparison of vector search using BAM embeddings with lexical search (BM25). Vector search is superior to lexical search, and improves on longer and more detailed queries (while lexical search degrades). }
  \label{fig:lexical}
\end{figure}

\paragraph{Comparison to Lexical Search}

Our production document retrieval service provides an ideal opportunity to benchmark the performance of vector search (using BAM embeddings) against traditional lexical search (using OpenSearch's Okapi BM25 implementation). To quantify the impact of query length, we benchmark queries containing 2--6 words, in each case using 1K randomly-selected company transcript queries from our test split.
Date ranges are restricted to a one year period that contains the positive passage. We evaluate both with and without filtering on the correct stock ticker.

As illustrated in Figure \ref{fig:lexical}, vector search outperforms lexical search in both settings, regardless of query length. While two-word queries are often ambiguous, vector search recall improves as queries become longer and more specific. In contrast, lexical search degrades on longer queries. This is consistent with our observation that users conditioned to using lexical search tools often limit their queries to 2 or 3 words.
To encourage users to write longer queries, in our frontend application we 
implement query completion/autocomplete using high-quality examples from our dataset.

\section{Conclusion}

We release text embeddings specialized for finance, trained on a dataset of 14.3M synthetic queries constructed from public and proprietary financial documents. On a held-out test set, BAM embeddings achieve 62.8\% Recall@1 vs. 39.2\% for the best general-purpose text embeddings. On the FinanceBench benchmark, replacing general-purpose embeddings with ours improves question answering accuracy by 8\%, demonstrating our dataset and model's ability to generalize to out-of-domain settings. Finally, we show that in a production document retrieval service, BAM embeddings outperform BM25 over all query lengths, and (unlike BM25) retrieval improves on longer and more detailed queries.

\newpage 

\bibliography{custom}

\newpage

\appendix

\section{Appendix}
\label{sec:appendix}

\subsection{Instructions to Query Annotators}
\label{sec:few-shot-instructions}

Your task is to read a text chunk and then write a query, such that the text chunk should be a top hit in a world-class retrieval system. 
Tips for writing queries:
\begin{enumerate}
    \item The query should be closely related to some part of the text (but not necessarily the entire passage).
    \item Queries can be from 1 to ~15 words.
    \item Queries can be a list of keywords, a full sentence, a question - whatever an analyst might search. Anything goes as long as the text chunk would be a very good search result for that query.
    \item Diversity is important, don't make your queries all the same. 
    \item Broad queries about entire industries, sectors, regions, or macro trends are fine, as long as the text snippet contains specific information that is highly relevant to the query.
    \item The query can mention a stock name or ticker. It doesn't have to.
    \item Don't copy too many words and phrases directly from the text passage. Use paraphrasing, synonyms, summarization, and your knowledge of appropriate abbreviations, acronyms and specialized terminology to construct queries. E.g., if the text mentions 'SBC', the query might mention `stock based comp'.
\end{enumerate}

\subsection{LLM Prompt for Query Generation}
\label{sec:query_prompt}

You are a highly trained investment analyst, and an expert in business and financial markets. You are helping construct a dataset to train a world class financial search engine. You will be given a text snippet from <DOCUMENT\_TYPE>. Your task is to generate a query derived from the provided text snippet.

\noindent
Detailed Instructions:
\begin{enumerate}
    \item The query should be a question, or a set of keywords or phrases, such that the text snippet should be returned as a top search result for that query.
    \item A good query is closely related to at least some, but not necessarily all, of the content in the text snippet. Do not create queries containing many unrelated concepts.
    \item Broad queries about entire industries, sectors, regions, or macro trends are okay, as long as the text snippet contains specific information that is relevant to the query. 
    \item You will be penalized if your query contains too many words and phrases copied directly from the text snippet. Use paraphrasing, synonyms, summarization, and your knowledge of appropriate abbreviations, acronyms and specialized terminology to construct queries. For example, if the text snippet contains the phrase "earnings per share", the query could instead include the acronym "EPS". If the text snippet mentions "earnings guidance for 2020-2024", the query could be for "long-term profit outlook".
    \item Do not include 10k or 10q references in your queries.
\end{enumerate}
 
\noindent
Formatting Instructions:
\begin{enumerate}
    \item Always reply with the query only, on a single line. Do not provide any additional context, note, or explanation of any kind. Do not put the query in quotation marks (",'). Do not include html tags in the query.
    \item Queries can contain a maximum of 10 words.
    \item If the text snippet is very short, difficult to understand, not written in English, or if it contains only boilerplate investment risk disclosures or disclaimers, you must begin your response with the special output "SKIP".
\end{enumerate}

\noindent
Text Snippet:
<EXAMPLE\_PASSAGE\_1>

\noindent
Query:
<EXAMPLE\_QUERY\_1>

\noindent
Text Snippet:
<EXAMPLE\_PASSAGE\_2>

\noindent
Query:
<EXAMPLE\_QUERY\_2>

\noindent
Text Snippet:
<SAMPLED\_PASSAGE>

\noindent
Query:

\end{document}